\documentclass[runningheads]{llncs}
\usepackage{graphicx}

\usepackage{tikz}
\usepackage{comment}
\usepackage{amsmath,amssymb} 
\usepackage{color}

\usepackage{graphicx}
\usepackage{amsmath}
\usepackage{amssymb}
\usepackage{booktabs}
\usepackage{adjustbox}
\usepackage{multirow}
\usepackage{marvosym}

\usepackage[accsupp]{axessibility}  

\begin{document}
\pagestyle{headings}
\mainmatter
\def\ECCVSubNumber{7694}  

\def\ie{\textit{i.e.}}
\def\eg{\textit{e.g.}}

\title{Reliability-Aware Prediction via Uncertainty Learning for Person Image Retrieval} 


\titlerunning{Reliability-Aware Prediction for Person Image Retrieval}
%
\author{Zhaopeng Dou\inst{1}, Zhongdao Wang\inst{1}, Weihua Chen\inst{2}, \\ Yali Li\inst{1}, \and Shengjin Wang~\inst{1}\textsuperscript{\Letter}
}

\authorrunning{Z. Dou et al.}

\institute{
Department of Electronic Engineering and BNRist, Tsinghua University, China \and
Machine Intelligence Technonlogy Lab, Alibaba Group\\
\email{dcp19@mails.tsinghua.edu.cn}, \email{wgsgj@tsinghua.edu.cn} 
}
\maketitle

\begin{abstract}
Current person image retrieval methods have achieved great improvements in accuracy metrics. 
However, they rarely describe the reliability of the prediction.
In this paper, we propose an Uncertainty-Aware Learning (UAL) method to remedy this issue. UAL aims at providing reliability-aware predictions by considering data uncertainty and model uncertainty simultaneously. Data uncertainty captures the ``noise" inherent in the sample, while model uncertainty depicts the model's confidence in the sample's prediction.
Specifically, in UAL, 
(1) we propose a sampling-free data uncertainty learning method to adaptively assign weights to different samples during training, down-weighting the low-quality ambiguous samples. 
(2) we leverage the Bayesian framework to model the model uncertainty by assuming the parameters of the network follow a Bernoulli distribution. 
(3) the data uncertainty and the model uncertainty are jointly learned in a unified network, and they serve as two fundamental criteria for the reliability assessment: if a probe is high-quality (low data uncertainty) and the model is confident in the prediction of the probe (low model uncertainty), the final ranking will be assessed as reliable. Experiments under the risk-controlled settings and the multi-query settings show the proposed reliability assessment is effective. 
Our method also shows superior performance on three challenging benchmarks under the vanilla single query settings. The code is available at: \textcolor{magenta}{\url{\textbf{https://github.com/dcp15/UAL}}}

\keywords{Person Image Retrieval, Uncertainty, Reliability Assessment.}
\end{abstract}

\footnote{The online version contains supplementary material available at \textcolor{magenta}{\url{\textbf{https://doi.org/10.1007/978-3-031-19781-9 34.}}}}

\section{Introduction}

\begin{figure}[t]
	\centering 
	\includegraphics[width=0.8\textwidth, height=0.25\textwidth]{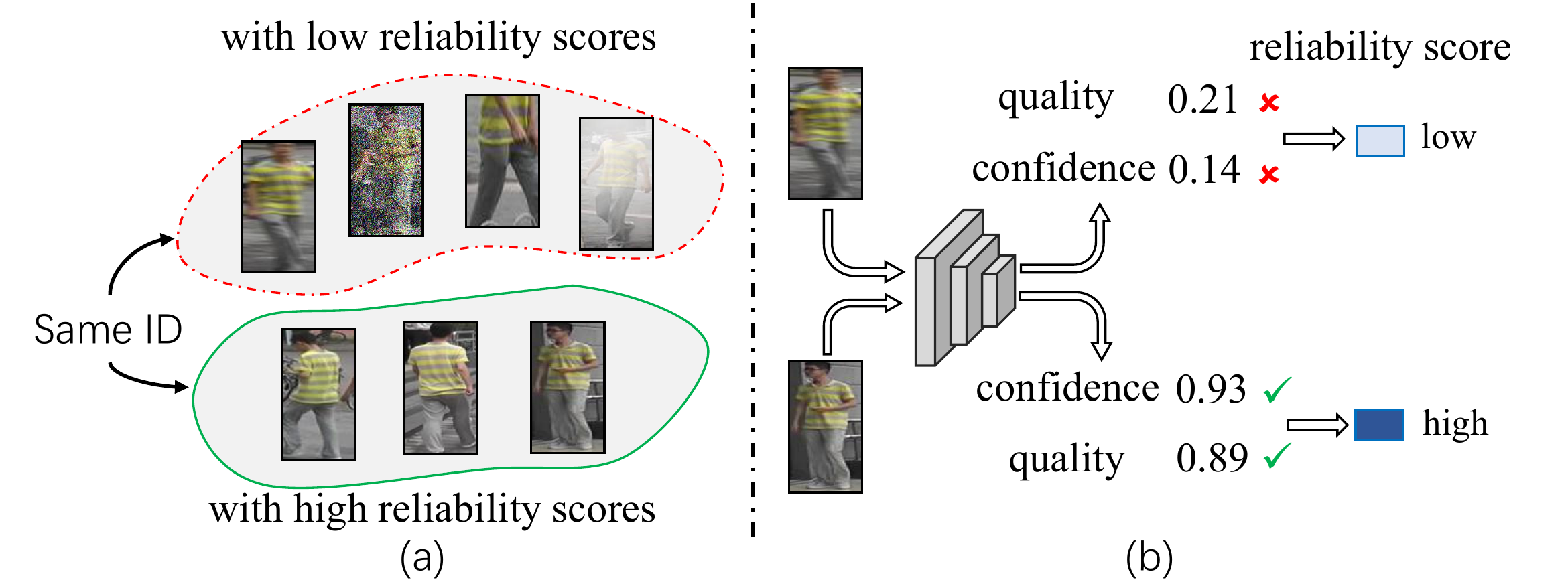}
	\caption{ Observation and Motivation.
	(a) In the multi-query setting, low-quality query images contain more ambiguous information. Reliability scores are required to down-weight the weights of these queries.
	(b) The reliability score is related to two factors: the quality of the sample and the model's confidence in the prediction of the sample.
	}
	\label{fig:introduction}
\end{figure}

Person image retrieval, also known as person re-identification (ReID), aims at associating a target person across non-overlapping camera views~\cite{PCB,MGN,ISP}. Although current methods~\cite{Mos,RFCnet,PAT,CDNet,VPM,transreid,IDM,isobe2021towards} have achieved promising performance on public benchmarks, they are reliability-agnostic, \ie, the prediction of a probe can be generated anyway, but they rarely describe whether the prediction is reliable. However, when people are identifying pedestrians, they not only give the judgment result but also the reliability associated with it. Such a reliability assessment mechanism is important in human decision-making~\cite{gawlikowski2021survey} and also essential in the ReID task. For example, in real scenarios, we often face the problem of searching for a person through his/her multiple images (\ie, multi-query settings), as a pedestrian is usually captured by several cameras and one camera may capture a series of observations of the person. More generally, we can add the retrieved positive ones into the query set for further comprehensive retrieval. The quality of these query images varies, especially in complex scenes. As shown in Fig.~\ref{fig:introduction}(a), low-quality query images contain more ambiguous information. If we treat these query images equally, performance will degrade. At this time, reliability scores are required to down-weight the low-quality query images. However, current methods rarely consider the reliability assessment problem.

To remedy this issue, we propose a novel Uncertainty-Aware Learning (UAL) method for the ReID task. UAL aims at not only giving an accurate prediction for a sample but also providing a reliability score associated with it. The reliability score is related to two factors, \ie, the quality of the sample and the confidence of the model in the prediction of the sample. These two factors are measured by considering two types of uncertainty, \ie, data uncertainty and model uncertainty. Data uncertainty captures the ``noise" inherent in the observation and it can describe the quality of the sample.
Model uncertainty represents the model's ``ignorance" and it can reflect the model's confidence in its prediction~\cite{kendall2017uncertainties,gal2016dropout}. 

In this paper, we propose a unified network to learn the data uncertainty and the model uncertainty simultaneously. Specifically, 
first, we project a sample into a Gaussian distribution in the latent space, the mean of the distribution represents the feature, and the variance represents the data uncertainty. Different from~\cite{DUL,Distribution_Net,PFE} sampling feature vector from the Gaussian distribution, we propose a sampling-free method to learn the data uncertainty and adaptively down-weight low-quality ambiguous samples during training.
Second, we leverage the Bayesian framework to learn the model uncertainty, in which the parameters of the network are assumed to follow the Bernoulli distribution. The model uncertainty is defined as the dispersion degree of the feature vectors caused by the distribution of the network parameter.
Third, the data uncertainty and model uncertainty are jointly learned in a unified network, and they serve as two criteria to assess whether the result is reliable: as shown in Fig.~\ref{fig:introduction}(b), if a query image is high-quality (low data uncertainty) and the model is confident in its prediction of the query image (low model uncertainty), the final result will be assessed as reliable. Experiments under risk-controlled settings and multi-query settings show the proposed reliability assessment is effective.

The major contributions are summarized as: 
\textbf{(1)} We propose an uncertainty-aware learning (UAL) method that can provide reliability-aware predictions for the ReID task.
\textbf{(2)} We introduce a sampling-free data uncertainty learning method, which can improve the representation by explicitly inhibiting the negative impact of low-quality samples during training without any external clues. 
\textbf{(3)} We propose a unified network to jointly learn data uncertainty and model uncertainty. As far as we know, this is the first work to apply data uncertainty and model uncertainty to the ReID task simultaneously.
\textbf{(4)} Experiments under risk-controlled settings and multi-query settings show the reliability assessment is effective. Our method also shows superior performance in single query settings. 

\section{Related Work}
\label{sec:related work}
\noindent\textbf{Person ReID.}  
Person ReID aims to associate a target person across different camera views. 
Existing methods can be broadly divided into two categories: hand-craft methods~\cite{liao2015person,yang2016large} and deep learning methods~\cite{li2021combined,huang2020real,isobe2021towards,IDM,transreid,Distribution_Net}. The key challenge is the large appearance variation caused by imperfect detection, different camera views, poses, and occlusions. To remedy these issues, several works~\cite{DSR,SFR,PGFA,HOReID,PVPM,ISP,FPR,PAT,Mos} are proposed to learn local features to cope with the appearance variation. Although these methods have played a certain role, they are reliability-agnostic. That is, the model can output a prediction for a probe anyway, but it does not describe the reliability of the prediction.

\textbf{Uncertainty in person ReID.} There are mainly two types of uncertainty: data uncertainty and model uncertainty~\cite{gal2016dropout,kendall2016modelling,shen2021real,kendall2017uncertainties}. Many tasks have considered the uncertainty to improve the robustness and interpretability of models, such as face recognition~\cite{PFE,khan2019striking,DUL}, semantic segmentation~\cite{isobe2017deep,kendall2017uncertainties} and Multi-view learning~\cite{geng2021uncertainty}. In the ReID task, prior arts~\cite{Distribution_Net,zheng2020exploiting,sun2021part,jin2020uncertainty} consider data uncertainty to alleviate the problem of label noise or data outliers. D-Net~\cite{Distribution_Net} maps each person image as a Gaussian distribution in the latent space with the variance indicating the data uncertainty. PUCNN~\cite{sun2021part} extends the data uncertainty in D-Net into the part-level feature. UNRN~\cite{zheng2020exploiting} incorporates the uncertainty into a teacher-student framework to evaluate the reliability of the predicted pseudo labels for unsupervised domain adaptive (UDA) person ReID. The uncertainty is estimated as the inconsistency of these two models in terms of their predicted soft multi-labels. UMTS~\cite{jin2020uncertainty} designs an uncertainty-aware knowledge distillation loss to transfer the knowledge of the multi-shots model into the single-shot model. Among these methods, the most relevant method to ours is D-Net~\cite{Distribution_Net}. Compared to D-Net, our data uncertainty learning method is sampling-free, which can explicitly suppress the ambiguous information contained in low-quality samples. We jointly learns the data uncertainty and the model uncertainty, which can utilize the complementary information provided by them during training. 

\section{Methodology}
\label{sec:proposed method}

The reliability score is related to two factors: the quality of the sample and the confidence of the model in its prediction, which are measured by data uncertainty (Sec.~\ref{subsec:Data uncertainty learning}) and model uncertainty (Sec.~\ref{subsec:model uncertainty learning}), respectively. They are incorporated into a unified network (Sec.~\ref{subsec: a unified network}) for joint learning. Two settings (risk-controlled and multi-query settings) are proposed to verify the effectiveness in Sec.~\ref{subsec:reliability assessment}.

\subsection{Learning Data Uncertainty}
\label{subsec:Data uncertainty learning}
Data uncertainty captures the ``noise" inherent in the observation. It can reflect the quality of the sample, which is an essential factor in the reliability assessment.

\textbf{Prior method.} Prior art D-Net~\cite{Distribution_Net} considers the data uncertainty by mapping a sample $\boldsymbol{x}$ as a Gaussian distribution in the latent space,
\begin{equation}
    p(\boldsymbol{z} | \boldsymbol{x}) = \mathcal{N}(\boldsymbol{z}; \boldsymbol{\mu}, \boldsymbol{\sigma}^2\mathbf{I})
    \label{eq: prior distribution}
\end{equation}
where $\boldsymbol{\mu}$ and $\boldsymbol{\sigma}^2$ are the mean and variance vectors. $\boldsymbol{\mu}$ is the feature vector and $\boldsymbol{\sigma}^2$ refers to the data uncertainty of $\boldsymbol{x}$. Then, they sample features from $p(\boldsymbol{z} | \boldsymbol{x})$ by re-parameterization trick~\cite{re_parameterization}: $\boldsymbol{z}' = \boldsymbol{\mu} + \boldsymbol{\epsilon} \boldsymbol{\sigma}, \boldsymbol{\epsilon}\sim \mathcal{N}(\mathbf{0}, \mathbf{I})$. The sampled $\boldsymbol{z}'$ are utilized for vanilla cross-entropy loss $\mathcal{L}_{ce}$. To prevent the trivial solution of variance decreasing to zero, a regularization term $\mathcal{L}_{fu}$ is added to constrain the entropy of $\mathcal{N}(\boldsymbol{\mu}, \boldsymbol{\sigma}^2\mathbf{I})$ to be larger than a constant. The final loss function is,
\begin{equation}
\mathcal{L} = \mathcal{L}_{ce} + \lambda \mathcal{L}_{fu}
\label{eq: prior art loss}
\end{equation}
where $\lambda$ is the hyper-parameter to balance $\mathcal{L}_{ce}$ and $\mathcal{L}_{fu}$. Although this method can capture the data uncertainty, there are two limitations: (1) it is sampling-based, \ie, the feature is sampled from the Gaussian distribution during training, which makes the optimization more difficult because each iteration optimizes only one point in the distribution, rather than entire distribution. (2) the objective does not explicitly distinguish samples with different data uncertainty. It is unclear how data uncertainty affects feature learning. To mitigate these two issues, we propose a sampling-free method to learn the data uncertainty and explicitly adjust the attention to the samples according to their quality.

\textbf{Our sampling-free data uncertainty learning method.} We project a sample $\boldsymbol{x}$ into a Gaussian distribution $\mathcal{N}(\boldsymbol{\mu}, \sigma^2\mathbf{I})$ in the latent space. Then the likelihood of $\boldsymbol{x}$ belonging to class $i$ is formulated by,
\begin{equation}
    p(\boldsymbol{x}|y=i) \propto \frac{1}{(2\pi\sigma^2)^{\frac{d}{2}}}\exp{(-\frac{\Vert\boldsymbol{\mu}-\boldsymbol{w}_i\Vert^2}{2\sigma^2})}
    \label{eq:likehood}
\end{equation}
where $\boldsymbol{w}_{i}$ is the weight vector of $i$-th class in the classifier and $d$ is the feature dimension. Assuming each class has the equal prior probability, the posterior of $\boldsymbol{x}$ belonging to the class $i$ is, 
\begin{equation}
    p(y=i|\boldsymbol{x}) = \frac{\exp{(-\frac{\Vert\boldsymbol{\mu}-\boldsymbol{w}_i\Vert^2}{2\sigma^2})}}{\sum\nolimits_{j}\exp(-\frac{\Vert\boldsymbol{\mu}-\boldsymbol{w}_j\Vert^2}{2\sigma^2})} = \frac{\exp{(\frac{1}{\sigma^2}\boldsymbol{w}_{i}^T\boldsymbol{\mu}})}{\sum\nolimits_{j}\exp{(\frac{1}{\sigma^2}\boldsymbol{w}_{j}^T\boldsymbol{\mu}})}
    \label{eq:posterior}
\end{equation}
Where $\boldsymbol{\mu}$ and $\boldsymbol{w}_*$ are $l_2$-normalized. $p(y|\boldsymbol{x})$ can be regarded as a Boltzmann distribution. The magnitude of $\sigma^2$ controls the entropy of this distribution. The larger the $\sigma^2$, the larger the entropy. Thus $\sigma^2$ can be regarded as the data uncertainty of sample $\boldsymbol{x}$. Assuming the class label of the sample $\boldsymbol{x}$ is $i$, the loss function is formulated by,
\begin{equation}
    \mathcal{L}_{d}(\boldsymbol{\mu}, \sigma^2) =  -\log{p(y=i|\boldsymbol{x})} \approx \frac{1}{\sigma^2}\mathcal{L}(\boldsymbol{\mu}) + \log\sigma^2
\label{eq:the final aleatoric uncertainty loss}
\end{equation}
where $\mathcal{L}(\boldsymbol{\mu})=-\log\frac{\exp{(\boldsymbol{w}_{i}^T\boldsymbol{\mu}})}{\sum\nolimits_{j}\exp{(\boldsymbol{w}_{j}^T\boldsymbol{\mu}})}$ is the cross entropy loss. Please see supplementary materials for derivation details.

\textbf{Discussion (difference to D-Net~\cite{Distribution_Net}).} Compared to sampling-based method D-Net~\cite{Distribution_Net}, our method shows several superior qualities: (1) Eq.~\ref{eq:the final aleatoric uncertainty loss} does not need to sample the representation from the Gaussian distribution. It contains entire information of the representation distribution. (2) Low-quality samples with larger data uncertainty will contribute less to learning the latent space. $\mathcal{L}(\boldsymbol{\mu})$ is weighted by $\frac{1}{\sigma^2}$, and thus it will drive the weight vector $\boldsymbol{w}$ in the classifier to be closer to high-quality samples with small $\sigma^2$. This can suppress the ambiguous information contained in low-quality samples during feature learning, which is verified in Sec.~\ref{subsec: analysis of the learned uncertainty}. (3) The $\sigma^2$ can not be too large or too small. If $\sigma^2$ is too small (large), the first (last) term $\frac{1}{\sigma^2}\mathcal{L}(\boldsymbol{\mu})$ ($\log\sigma^2$) becomes too large. Unlike the term $\mathcal{L}_{fu}$ in Eq.~\ref{eq: prior art loss}, we maintain the $\sigma^2$ in a unified formulation.

\subsection{Learning Model Uncertainty}
\label{subsec:model uncertainty learning}
Model uncertainty plays an important role in the reliability assessment as it reflects the confidence of the model in its prediction of a sample.
Bayesian network is often used to capture the model uncertainty~\cite{kendall2017uncertainties}. Suppose we have a dataset $\mathcal{D}=(\boldsymbol{X,Y})$, we define $f_{\boldsymbol{\theta}}$ to be a neural network such that $f_{\boldsymbol{\theta}}:\boldsymbol{X}\rightarrow\boldsymbol{Y}$ and $\boldsymbol{\theta}$ corresponds to the weight of the network. To capture model uncertainty, we assume a prior distribution on the weight, \ie, $p(\boldsymbol{\theta})$. We need to obtain the posterior $p(\boldsymbol{\theta}|\mathcal{D})$. Since $p(\boldsymbol{\theta}|\mathcal{D})$ is typically intractable, an approximate distribution $q_{\boldsymbol{\pi}}(\boldsymbol{\theta})$ parameterized by $\boldsymbol{\pi}$ is defined.  $q_{\boldsymbol{\pi}}(\boldsymbol{\theta})$ is aimed to be as similar as possible to $p(\boldsymbol{\theta}|\mathcal{D})$, which is measured by the Kullback-Leibler divergence. The optimal parameters $\boldsymbol{\pi}^{*}$ is,
\begin{equation}
\begin{aligned}
    \boldsymbol{\pi}^{*} = \mathop{\arg\min}\limits_{\boldsymbol{\pi}} \mathrm{KL}[q_{\boldsymbol{\pi}}(\boldsymbol{\theta}) \Vert p(\boldsymbol{\theta}|\mathcal{D})] 
    = \mathop{\arg\min}\limits_{\boldsymbol{\pi}} \mathrm{KL}[q_{\boldsymbol{\pi}}(\boldsymbol{\theta})\Vert p(\boldsymbol{\theta})] - \mathbb{E}_{q_{\boldsymbol{\pi}}(\boldsymbol{\theta})}[\log{p(\mathcal{D}|\boldsymbol{\theta})}] \\
\end{aligned}
\label{eq:bayes variational inference}
\end{equation}
Please see supplementary materials for derivation details. 
Similar to~\cite{gal2015bayesian},  we assume the prior distribution $p(\boldsymbol{\theta})$ as the Bernoulli distribution and perform Monte Carlo integration to the second term. Then the objective loss can be written as $\mathcal{L}_{m} = -\frac{1}{T}\sum_{t=1}^{T}[\log{p(\mathcal{D}|\boldsymbol{\theta}_{t})}]$, where $\boldsymbol{\theta}_{t}$ is sampled from $q_{\boldsymbol{\pi}}(\boldsymbol{\theta})$. In $q_{\boldsymbol{\pi}}(\boldsymbol{\theta})$, $\boldsymbol{\theta}_{ij} = \boldsymbol{\pi}_{ij} * \boldsymbol{z}_{ij}$, where $\boldsymbol{z}_{ij}\sim \mathrm{Bernoulli}(\rho)$. $\rho$ is the hyper-parameters and $\boldsymbol{\pi}$ is the set of parameters to be optimized. For a sample $\boldsymbol{x}$, we denote the extracted feature is $\boldsymbol{\mu}_{t} = f_{\boldsymbol{\theta}_{t}}(\boldsymbol{x})$ when the weight of network is $\boldsymbol{\theta}_{t}$. The model uncertainty is defined by the variance of the features over the network parameter distribution,
\begin{equation}
\boldsymbol{\sigma}^2_m = \frac{1}{T}\sum_{t=1}^{T}(\boldsymbol{\mu}_{t}-\bar{\boldsymbol{\mu}})^2
\label{eq: model_uncertainty}
\end{equation}
where $\bar{\boldsymbol{\mu}}  = \frac{1}{T}\sum_{t=1}^{T} \boldsymbol{\mu}_{t}$ and squaring operations in Eq.~\ref{eq: model_uncertainty} are element-wise. 

\begin{figure}[t]
	\centering 
	\includegraphics[width=0.85\linewidth]{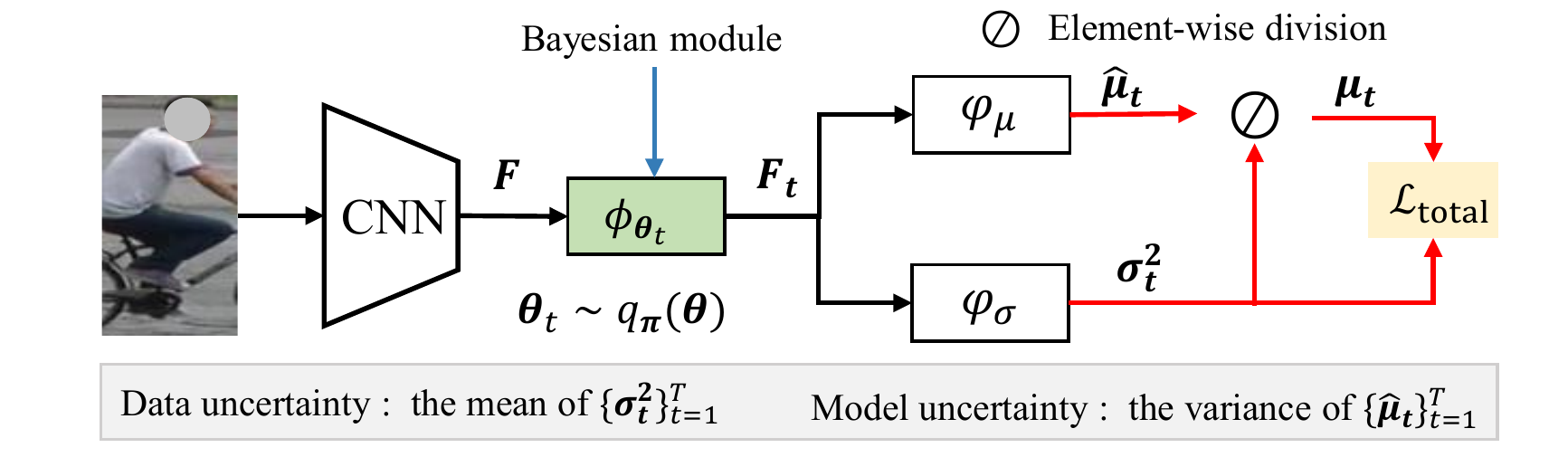}
	\caption{Overview of the proposed method. An image is fed into the backbone (CNN) to obtain the feature maps $\boldsymbol{F}$. $\boldsymbol{F}$ is further input into the Bayesian module $\phi$ whose parameter is $\boldsymbol{\theta}_{t}$ to obtain $\boldsymbol{F}_{t}$. Then $\boldsymbol{F}_{t}$ is input to two deterministic modules $\varphi_{\mu}$ and $\varphi_{\sigma}$ to predict the feature $\hat{\boldsymbol{\mu}}_{t}$, $\boldsymbol{\mu}_{t}$ and the data uncertainty $\boldsymbol{\sigma}^2_{t}$, respectively. 
	The model uncertainty is the variance of $\{\hat{\boldsymbol{\mu}}_{t}\}_{t=1}^{T}$, and the data uncertainty is the mean of $\{\boldsymbol{\sigma}^2_{t}\}_{t=1}^{T}$.
	}
	\label{fig:overview}
\end{figure}

\subsection{Jointly Learning Data and Model Uncertainty}
\label{subsec: a unified network}

To avoid the additional overhead caused by separately learning the data uncertainty and the model uncertainty, and at the same time leverage the complementary information provided by them for representation learning, we integrate them into a unified network for joint learning.
As shown in Fig.~\ref{fig:overview}, given a sample $\boldsymbol{x}$, we first feed it into a CNN backbone to get the feature maps $\boldsymbol{F}\in \mathbb{R}^{h\times w\times c}$, where $h$, $w$, $c$ are height, weight and channel, respectively. $\boldsymbol{F}$ is further fed into a Bayesian module $\phi$ to obtain $\boldsymbol{F}_{t} = \phi_{\boldsymbol{\theta}_{t}}(\boldsymbol{F})$, where $\boldsymbol{\theta}_{t}\sim q_{\boldsymbol{\pi}}(\boldsymbol{\theta})$ corresponds to the parameters of $\phi$. Then we input $\boldsymbol{F}_{t}$ into two deterministic modules $\varphi_{\mu}$ and $\varphi_{\sigma}$ to obtain the embedding $\hat{\boldsymbol{\mu}}_{t} = \varphi_{\mu}(\boldsymbol{F}_{t})$ and data uncertainty $\boldsymbol{\sigma}^2_{t}=\varphi_{\sigma}(\boldsymbol{F}_{t})$, respectively. Here, $\boldsymbol{F}_{t}$, $\hat{\boldsymbol{\mu}}_{t}$, $\boldsymbol{\sigma}^2_{t}$ have the same shape as $\boldsymbol{F}$.
We further regularize $\hat{\boldsymbol{\mu}}_{t}$ by quality-aware pooling: $\boldsymbol{\mu}_{t} =\text{GAP}(\boldsymbol{\hat{\mu}}_{t}\oslash{\boldsymbol{\sigma}^2_t}) \in \mathbb{R}^{c}$, where $\oslash$ represents the element-wise division and GAP refers to the global average pooling. Thus, for the sample $\boldsymbol{x}$, we get its feature distribution  $\mathcal{N}(\boldsymbol{\mu}_{t}, \sigma^2_{t}\mathbf{I})$, where $\sigma^2_{t}$ is the mean of $\boldsymbol{\sigma}^2_{t}$ across all elements. 
During training, the loss function is,
\begin{equation}
\mathcal{L}_{\mathrm{total}} = \mathcal{L}_{d}(\boldsymbol{\mu}_t, \sigma^2_t) + \mathcal{L}_{\mathrm{tri}}(\boldsymbol{\mu}_{t})
\label{eq: total loss}
\end{equation}
where $\mathcal{L}_{d}$ is defined in Eq.~\ref{eq:the final aleatoric uncertainty loss} and $\mathcal{L}_{\mathrm{tri}}(\boldsymbol{\mu}_{t})$ is the triplet loss~\cite{triplet_loss} on $\boldsymbol{\mu}_{t}$.

In inference, for a sample $\boldsymbol{x}$, we can sample $T$ times from $q_{\boldsymbol{\pi}}(\boldsymbol{\theta})$ and obtain $\{\hat{\boldsymbol{\mu}}_{t}\}_{t=1}^{T}$, $\{\boldsymbol{\mu}_{t}\}_{t=1}^{T}$ and $\{\sigma^2_{t}\}_{t=1}^{T}$. The data uncertainty is formulated by $\sigma^2_{d}=\frac{1}{T}\sum_{t=1}^T\sigma^2_{t}$. Then $\boldsymbol{\sigma}^2_{m}$ is estimated as the variance of $\{\hat{\boldsymbol{\mu}}_{t}\}_{t=1}^{T}$ according to Eq.~\ref{eq: model_uncertainty} and the model uncertainty $\sigma^2_m$ is defined as the mean of $\boldsymbol{\sigma}^{2}_{m}$ across all elements. The final representation is calculated by $\bar{\boldsymbol{\mu}}=\frac{1}{T}\sum_{t=1}^{T}\boldsymbol{\mu}_{t}$.

Note that $\boldsymbol{\sigma}^2_{t}$ is the data uncertainty of $\boldsymbol{x}$ under the parameter $\boldsymbol{\theta}_{t}$. In practice, we train the network to predict $\boldsymbol{s}_{t}$, and $\boldsymbol{\sigma}^2_t:=\mathrm{log}(1 + \mathrm{exp}(\boldsymbol{s}_{t}))$. This is because that it is more numerically stable than directly predicting the $\boldsymbol{\sigma}^2_{t}$, as it ensures that each element of $\boldsymbol{\sigma}^2_{t}$ is greater than zero. $\phi$, $\varphi_{\mu}$ and $\varphi_{\sigma}$ are all light weighted modules and their architectures are detailed in supplementary materials.

\subsection{Reliability Assessment}
\label{subsec:reliability assessment}
Based on the risk-controlled settings and the multi-query settings, we introduce how to leverage the learned data and model uncertainty for reliability assessment.

\textbf{Risk-controlled settings.} When facing complex application scenarios, to control the cost of errors, we would expect the model to reject input images (probes) if it can not deal with them. We show the proposed uncertainty mechanism can be naturally utilized as such a ``risk indicator". Given a probe $\boldsymbol{x}$, we can obtain its data uncertainty $\sigma^2_d$ and model uncertainty $\sigma^2_m$. The probe will be assessed as safety (the prediction is reliable) only if $\frac{1}{\sigma^2_d} > \gamma_{d}$ and $\frac{1}{\sigma^2_m} > \gamma_{m}$, where $\gamma_{d}$ and $\gamma_{m}$ are two thresholds that can be set based on the risk tolerance.

\textbf{Multi-query settings.} In real scenarios, how to utilize multiple query images from the same identity to search this person is an essential issue as an interested pedestrian is usually captured by several cameras. Our method is naturally suitable for such settings because it can suppress the negative impact of ambiguous queries according to the reliability score. Considering $\mathcal{X}=\{\boldsymbol{x}_{1}, \dots, \boldsymbol{x}_{n}\}$ are query images from the same identity and $\boldsymbol{y}$ is an image in the gallery set. The key issue is how to measure the similarity between $\mathcal{X}$ and $\boldsymbol{y}$.
Let $\sigma^2_{d, i}$ ($\sigma^2_{m, i}$) be the data (model) uncertainty of $\boldsymbol{x}_{i}$. To combine the data uncertainty and model uncertainty without being affected by their numerical scale, we project the data (model) uncertainty into the interval $[\tau_{\text{min}}, \tau_{\text{max}}]$. Specifically, $\sigma^2_{d,i}$ is projected to $d_{i}=\beta_{i}\tau_{\text{min}} + (1 - \beta_{i})\tau_{\text{max}}$, where $\beta_{i}=\frac{\sigma^2_{d,i}-\sigma^2_{d, \text{min}}}{\sigma^2_{d,\text{max}} - \sigma^2_{d, \text{min}}}$ and $\sigma^2_{d,\text{max}}$ ($\sigma^2_{d, \text{min}}$) is the maximum (minimum) value in $\{\sigma^2_{d, i}\}_{i=1}^{n}$. The model uncertainty $\sigma^2_{m, i}$ is mapped to $m_{i}$ in the similar way. The \textbf{reliability score} of $\boldsymbol{x}_i$ is $w_{i}=\frac{m_{i}d_{i}}{\sum_{i=1}^{n}m_{j}d_{j}}$. The similarity between $\mathcal{X}$ and $\boldsymbol{y}$ is calculated by $s=\sum_{i=1}^{n}w_{i}s_{i}$, where $s_{i}$ is the similarity between $\boldsymbol{x}_{i}$ and $\boldsymbol{y}$. The similarity $s$ considers the reliability of each element in $\mathcal{X}$. If an element has larger reliability score, it plays a more important role.

\section{Experiments}
\label{sec:experiments}
\subsection{Datasets and Evaluation Metrics}
\label{subsec: datasets and evaluation metrics}

The datasets we use include Market-1501~\cite{market}, MSMT17~\cite{MSMT17}, CUHK03~\cite{CUHK03,CUHK03NP}, Occluded-Duke~\cite{Occluded-Duke}, Occluded-REID~\cite{Occluded-REID} and Partial-REID~\cite{Partial-REID}. \textbf{Market-1501} has 12,936 training, 3,368 query and 19,732 gallery images. \textbf{CUHK03} contains 13,164 images of 1,467 identities. We adopt the new testing protocol proposed in~\cite{CUHK03NP}. \textbf{MSMT17} is the largest image dataset for person ReID. It contains 126,441 images of 4,101 identities. \textbf{Occluded-Duke} is reconstructed from DukeMTMC-reID~\cite{Duke} by selecting occluded images as query set. It has 15,618 training images, 2,210 query and 17,661 gallery images. \textbf{Occluded-REID} contains 2,000 images belonging to 200 identities. Each identity has five occluded person images and five-full-body images. \textbf{Partial-REID} contains 600 images from 60 person, with five partial images and five full-body images per person. 

\textbf{Evaluation metrics.} We use the Cumulative Matching Characteristic (CMC) curve and mean average precision (mAP) as the evaluation metric.

\subsection{Implementation Details}
\label{subsec: implementation details}

Input images are resized to $256\times 128$. During training, images are augmented by random cropping, random horizontal flipping, and random erasing~\cite{random_earsing}. Following~\cite{PCB}, we adopt ResNet50~\cite{ResNet} pre-trained on ImageNet as the backbone. For a fair comparison with methods employing more complex backbone networks, following ISP~\cite{ISP}, we also employ HRNet-W32~\cite{hrnet} as our backbone. There are 64 images from 16 identities in a mini-batch. The initial learning rate is $3.5\times10^{-4}$. For MSMT17~\cite{MSMT17} (other datasets), we train our model 160 (120) epochs, and the learning rate is decreased to its 0.1 and 0.01 at the 70$^{\mathrm{th}}$ (40$^{\mathrm{th}}$) and 110$^{\mathrm{th}}$ (70$^{\mathrm{th}}$) epochs, respectively. Each epoch has 200 iterations. The optimizer is Adam~\cite{adam}. The $\rho$ of Bernoulli distribution in the Bayesian module $\phi$ is empirically set as 0.7. The $\tau_{\text{min}}$ and $\tau_{\text{max}}$ in the multi-query settings are set as 0.5 and 1.0, respectively. 

During testing, we first obtain the parameters $\{\boldsymbol{\theta}_{t}\}_{t=1}^{T}$ of the Bayesian module $\phi$ by sampling $T$ times from $q_{\boldsymbol{\pi}}(\boldsymbol{\theta})$. For each $\boldsymbol{\theta}_{t}$, we use it to handle all samples. This ensures that the features of different samples are extracted by the same network, which makes it possible to employ our method with a small $T$. As the Bayesian module $\phi$ is at the back of the entire network, we only need to repeatedly input it into $\phi$, $\varphi_{\mu}$ and $\varphi_{\sigma}$ for $T$ times, rather than entire network, which saves a lot of expenditure (related results are shown in Table~\ref{table:ablation of T.}). 

\subsection{Experiments on the Reliability Assessment}
\label{subsec: experiments on reliability assessment}

\begin{figure}[t]
	\centering 
	\includegraphics[width=1.0\linewidth]{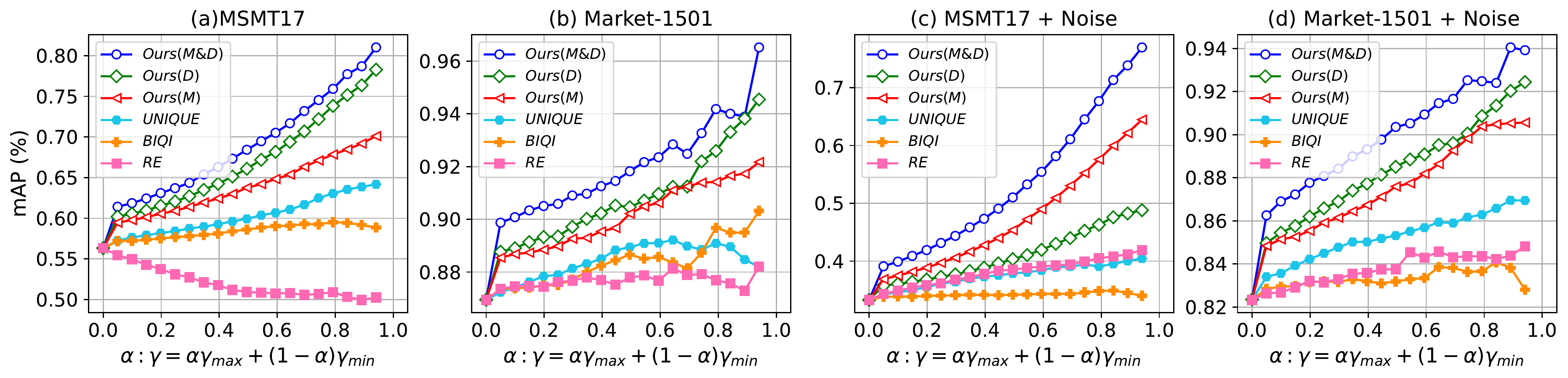}
	\caption{ Comparison of the risk-controlled settings. Filtering out a proportion of query images according to the thresholds of different criteria. \textit{D}: data uncertainty; \textit{M}: model uncertainty; \textit{M$\&$D}: model uncertainty and data uncertainty; \textit{BIQI}~\cite{BIQI}: blind image quality indices. \textit{RE}: the reciprocal of the entropy of the predicted category distribution. \textit{UNIQUE}~\cite{UNIQUE}: unified no-reference image quality and uncertainty evaluator.
	}
	\label{fig:risk_control}
\end{figure}

\noindent\textbf{Risk-controlled settings.} Here, we design experiments to verify whether the proposed uncertainty can serves as the ``risk indicator" as described in Sec.~\ref{subsec:reliability assessment}. Specifically, we allow the model to filter out some queries it is diffident to maintain higher performance. Here, we consider four criteria for filtering, \ie, 
(1) \textit{UNIQUE}~\cite{UNIQUE}: unified no-reference image quality and uncertainty evaluator;
(2) \textit{BIQI}~\cite{BIQI}: blind image quality indices;
(3) \textit{RE}: the reciprocal of the entropy of the predicted category distribution;
(4) \textit{Ours}: the reciprocal of the proposed uncertainty as described in Sec.~\ref{subsec:reliability assessment}. (\textit{M:} model uncertainty alone; \textit{D:} data uncertainty alone; \textit{M$\&$D:} combing model uncertainty and data uncertainty). For each criterion, we first calculate its maximum and minimum on all queries, denoted as $\gamma_{\text{max}}$ and $\gamma_{\text{min}}$. Then we setting the threshold as $\gamma = \alpha\gamma_{\text{max}} + (1 - \alpha)\gamma_{\text{min}}$. For \textit{M$\&$D}, a query will be kept only if $\frac{1}{\sigma^2_d} > \gamma_{d,\alpha}$ and $\frac{1}{\sigma^2_m} > \gamma_{m, \alpha}$, where $\sigma^2_d$ and $\sigma^2_m$ are the data uncertainty and the model uncertainty, and $\gamma_{d, \alpha}$ and $\gamma_{m, \alpha}$ are their thresholds at parameter $\alpha$.   
We report the mAP score against the $\alpha$ under different settings. Fig.~\ref{fig:risk_control}(a-b) shows the results on MSMT17 and Market-1501. In Fig.~\ref{fig:risk_control}(c-d), we add Gaussian noise to the images in the query set with a probability of 0.5.
From the results, we can make several observations. 
First, for our methods (\textit{M}, \textit{D}, and \textit{M$\&$D}), as $\alpha$ increasing, the mAP score of the remained queries increases. This shows that the proposed uncertainty mechanism help filter out queries whose prediction is unreliable (the model can not deal with) while retaining queries the model is capable of. 
Second, under different settings, model uncertainty or data uncertainty alone already outperforms \textit{UNIQUE}, \textit{BIQI} and \textit{RE}. When they are combined, the mAP score of retained queries is higher. This shows the proposed uncertainty mechanism is a better indicator of the reliability of the prediction, and thus helps for making risk-controlled decisions.

\label{subsec: multi-queries experiments.}
\begin{table}[t]
\begin{center}
\caption{
Results of the multi-query settings on Market-1501~\cite{market} and MSMT17~\cite{MSMT17} and Occluded-Duke~\cite{Occluded-Duke}. To simulate the complex scenes, we transform the query images via different quality-degrading transformations.
MT17$\rightarrow$Market: the model is trained on MSMT17 and directly tested on Market-1501. ``\textbf{\textit{w/}} R" (``\textbf{\textit{w/o}} R") means we use (don't use) the reliability score to adjust the weights for different queries.
} 
\label{table:results about mutiple queries.}
\begin{tabular}{c|c|cc|cc|cc|cc}
\toprule
\multirow{2}{*}{Complex scenes} &   \multirow{2}{*}{Method} & \multicolumn{2}{c|}{Market-1501} & \multicolumn{2}{c|}{MSMT17} & \multicolumn{2}{c|}{Occlu-Duke}& \multicolumn{2}{c}{MT17$\rightarrow$Market} \\ \cline{3-10}
                        &    & R1   & mAP   &R1  &mAP       & R1          & mAP     & R1    &mAP  \\  \hline
\multirow{3}{*}{Gaussian Noise} 
                                
                                & D-Net\cite{Distribution_Net}      &  54.5      & 52.6        &  39.8       & 26.5       &  43.8      & 35.9     & 25.0   & 12.9 \\
                                & Ours (\textbf{\textit{w/o}} R)   &  55.0     &  53.1      & 49.9          & 34.5    & 51.3    & 43.7     & 32.2   & 18.3   \\  
                                & \textbf{Ours} (\textbf{\textit{w/}} R)    &\textbf{76.7}     &  \textbf{71.4}      & \textbf{62.9}          & \textbf{45.4}  &  \textbf{61.5}     & \textbf{53.8}       &\textbf{36.8}    & \textbf{20.6}    \\  \hline
\multirow{3}{*}{Random Crop}   
                              
                              & D-Net\cite{Distribution_Net}    & 36.5 &  33.6 &  52.6   & 33.7    & 23.8   & 20.3  & 22.5   &  12.7 \\
                              & Ours (\textbf{\textit{w/o}} R)  &  37.4     &  36.0      & 58.4          & 38.6   & 27.0   & 23.4   & 28.6   & 16.4   \\
                              & \textbf{Ours} (\textbf{\textit{w/}} R)    &  \textbf{46.3}     &  \textbf{44.4}      & \textbf{63.1}          & \textbf{42.8}  & \textbf{35.1}    & \textbf{31.3}   &\textbf{32.7}    & \textbf{18.8}    \\  \hline
\multirow{3}{*}{Motion Blur}  
                              & D-Net\cite{Distribution_Net} & 66.9 & 59.3 & 55.9  & 38.0 & 60.0  & 50.2   & 25.5   & 14.2 \\
                              & Ours (\textbf{\textit{w/o}} R)  &  69.7     &  63.4      &  66.7        & 47.1   & 62.0   & 53.3    & 35.5  & 20.8    \\
                              & \textbf{Ours} (\textbf{\textit{w/}} R)  & \textbf{77.4}     &  \textbf{71.4}      & \textbf{75.3}          & \textbf{54.9}  & \textbf{66.4}   & \textbf{58.5}   & \textbf{46.3}    &  \textbf{26.3}    \\  \hline
\multirow{3}{*}{Adding fog}  
                              & D-Net\cite{Distribution_Net}  &  62.1    & 56.6  & 50.9  & 33.7   & 50.3  & 42.1  & 29.0  & 15.7 \\
                              & Ours (\textbf{\textit{w/o}} R) &  65.1     &  60.3      & 58.3         & 40.2    &  55.9  & 47.5   & 37.4         & 21.9     \\
                              & \textbf{Ours} (\textbf{\textit{w/}} R)  & \textbf{77.1}     &  \textbf{71.9}      & \textbf{72.0}         & \textbf{52.0}  &  \textbf{64.1}   & \textbf{56.6}    & \textbf{47.0}    &  \textbf{26.9}    \\  
\bottomrule
\end{tabular}
\end{center}
\end{table}

\begin{table}[t]
\begin{center}
\caption{
Results of multi-query settings on Occluded-REID~\cite{Occluded-REID}, Occluded-Duke~\cite{Occluded-Duke} and Partial-REID~\cite{Partial-REID}. The query images are \textbf{not} transformed to degrade the quality.
}
\label{table:results about mutiple queries on occluded dataset.}
\begin{tabular}{c|cc|cc|cc}
\toprule
\multirow{2}{*}{Method} & \multicolumn{2}{c|}{Occluded-Duke}&\multicolumn{2}{c|}{Occluded-REID} & \multicolumn{2}{c}{Partial-REID}  \\ \cline{2-7}
                        & R1   & mAP  & R1   & mAP         & R1          & mAP  \\  \hline

D-Net\cite{Distribution_Net} & 75.9   & 67.0   & 82.0   & 72.3   & 80.0    & 74.4    \\
Ours (\textbf{\textit{w/o}} R)  & 76.6 & 68.3 & 80.0     &  72.2      & 85.0        & 77.1   \\   
\textbf{Ours} (\textbf{\textit{w/}} R) & \textbf{78.0} & \textbf{70.5} & \textbf{83.0}     &  \textbf{73.5}      & \textbf{88.3}    & \textbf{80.0}    \\  
\bottomrule
\end{tabular}
\end{center}

\end{table}

\begin{figure}[t]
	\centering 
	\includegraphics[width=0.9\linewidth, height=0.15\linewidth]{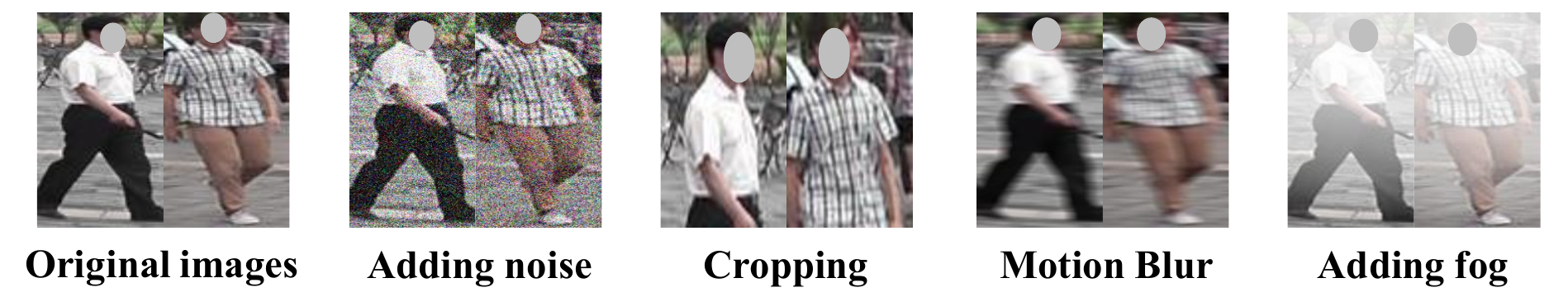}
	\caption{ Visualization of different quality-degrading transformations.}
	\label{fig:figure_examples}
\end{figure}

\textbf{Multi-query settings.} We find that there are few images with the same personal identity and same camera identity in the query set of existing datasets. Thus, we reconstruct the test sets of Market-1501, Occluded-Duke and MSMT17 to evaluate the proposed reliability assessment under multi-query settings. Specifically, for each dataset, we first collect the images belonging to the same personal identity and same camera identity from the query set and gallery set. Then, we randomly select half of these images to be allocated to the reorganized query set and the other half to the reorganized gallery set. Simultaneously, to simulate the complex scenes in reality, we transform images in the reorganized query set with a certain probability. The transformation includes: (1) add Gaussian noise to mimic camera quality differences; (2) crop the image to simulate pedestrians are partially out of the camera's field of view; (3) add motion blur to mimic fast-moving pedestrians; (4) add fog to imitate complex weather. Please refer to the supplementary materials for the transformation details. Fig.~\ref{fig:figure_examples} shows some examples of these transformations.  
In inference, the images belonging to the same personal identity and same camera identity in the query set are regarded as a set of templates that need to be associated.
We report the performance of our method against the D-Net~\cite{Distribution_Net} in Table.~\ref{table:results about mutiple queries.}. ``\textbf{\textit{w/}} R" (``\textbf{\textit{w/o}} R") means we use (don't use) the reliability score to adjust the weights for different queries. From the results, we can make several observations.
First, for different complex scenes and different datasets, our method (``\textbf{\textit{w/o}} R") already outperforms D-Net~\cite{Distribution_Net}. This shows that our method learns better embedding space, in which the features are more discriminative. When we additionally use the reliability score to adjust the weights for different queries, the performance is further improved. This shows that the proposed reliability score is credible, which can help mine more valuable queries and suppress distractions from low-quality ones.

We also evaluate our method on Occluded-REID~\cite{Occluded-REID}, Occluded-Duke~\cite{Occluded-Duke} and Partial-REID~\cite{Partial-REID}. As query images in these three datasets are occluded or partial, we directly test on them without any quality degradation transformation. Results are shown in Table~\ref{table:results about mutiple queries on occluded dataset.}. We can draw the same conclusions as in Table~\ref{table:results about mutiple queries.}.

\subsection{Analysis of the Reliability Assessment.}
\label{subsec: analysis of the learned uncertainty}

\begin{figure}[t]
\begin{minipage}[h]{0.49\textwidth}

\includegraphics[width=1.0\textwidth, height=0.5\textwidth]{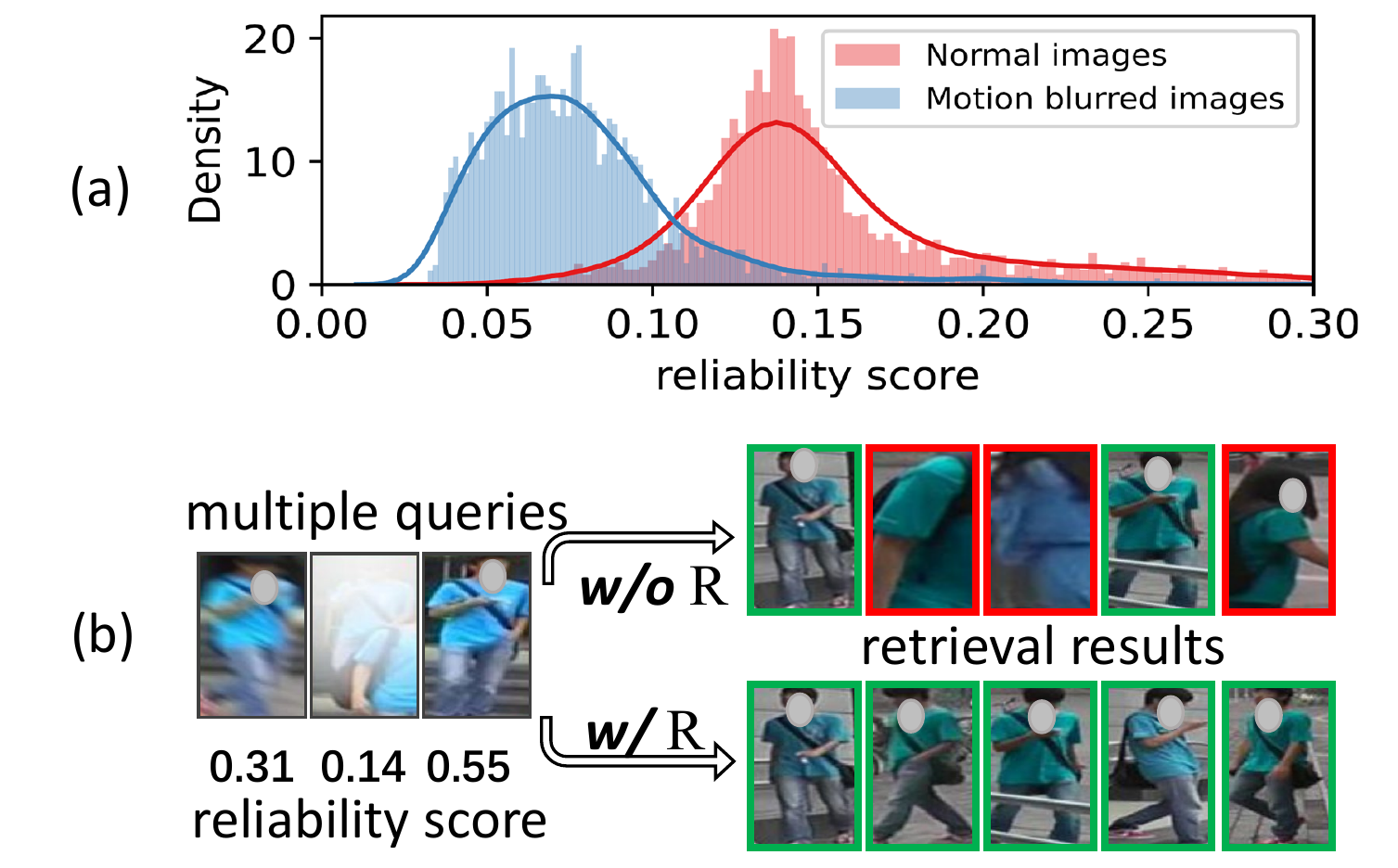}
\caption{ (a) Distributions of the reliability scores under multi-query settings. (Normal images $v.s.$ Motion-blurred images).
(b) An example. ``\textbf{\textit{w/o}} R" (``\textbf{\textit{w/}} R") means we use (don't use) the reliability score to adjust the weights for different queries.
}
\label{fig:multi-query-visualization}
\end{minipage}
\hfill
\begin{minipage}[h]{0.49\textwidth}
	\includegraphics[width=1.0\textwidth, height=0.49\textwidth]{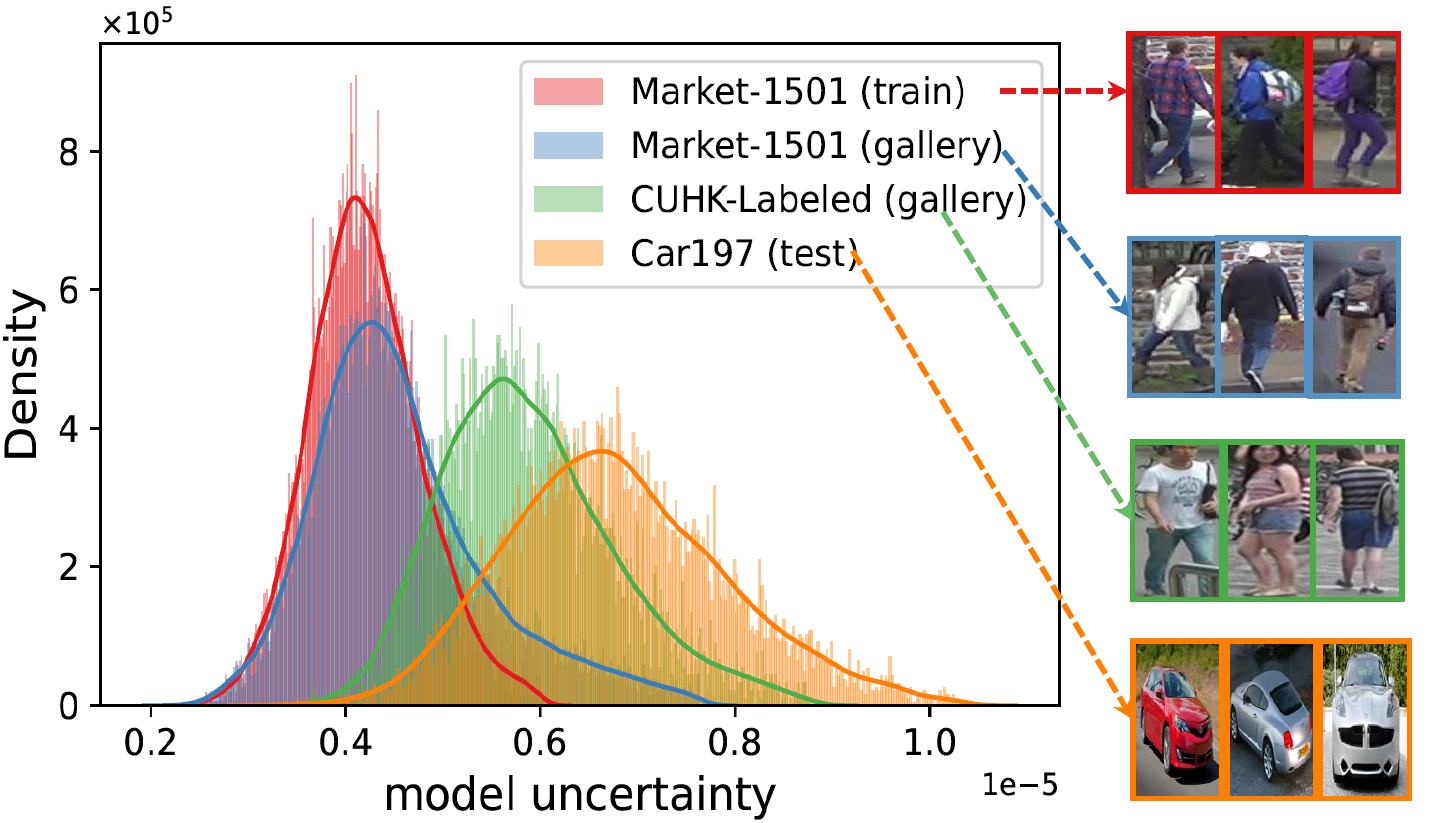}
	\caption{The role of model uncertainty in reliability assessment. The model is trained with the training set of the Market-1501. The learned model uncertainty is proportional to the degree of deviation between test set domain and training set domain.
	}
	\label{fig:model_uncertainty_capture}
\end{minipage}
\end{figure}

\noindent\textbf{Is the reliability score reasonable?} Here, we analyze the reasonableness of the reliability score based on the multi-query settings. For each identity in the test set of Market-1501, we randomly select 10 images to form multiple queries, in which 5 images are downgraded in quality by adding motion blur. Fig.~\ref{fig:multi-query-visualization} (a) shows the distributions of reliability scores for normal images and motion-blurred images. We can see that, on average, the reliability score of normal images is greater than that of motion-blurred images, showing the reliability score is credible. The retrieval example in Fig.~\ref{fig:multi-query-visualization} (b) shows that when we adjust the weights of different queries according to the reliability scores, the ambiguous information from the low-quality query is suppressed, resulting in a better performance. 

\textbf{The role of model uncertainty in reliability assessment.} We design experiments to verify whether the proposed model uncertainty can describe the confidence of the model in its prediction of the sample. We use the training set of Market-1501 to train the model. Then we estimate the model uncertainty of samples from the training set of Market-1501, the gallery set of Market-1501, the gallery set of CUHK03 and the test set of Car197~\cite{car197}, respectively. Fig.~\ref{fig:model_uncertainty_capture} shows the results. On average, 
with the increase of the domain deviation, the model uncertainty gradually grows. This shows that the estimated model uncertainty is related to the model's prediction confidence. For out-of-distribution inputs, the model is diffident about the predictions, and the model uncertainty is large.

\begin{figure}[t]
\begin{minipage}[h]{0.49\textwidth}
\includegraphics[width=1.0\textwidth]{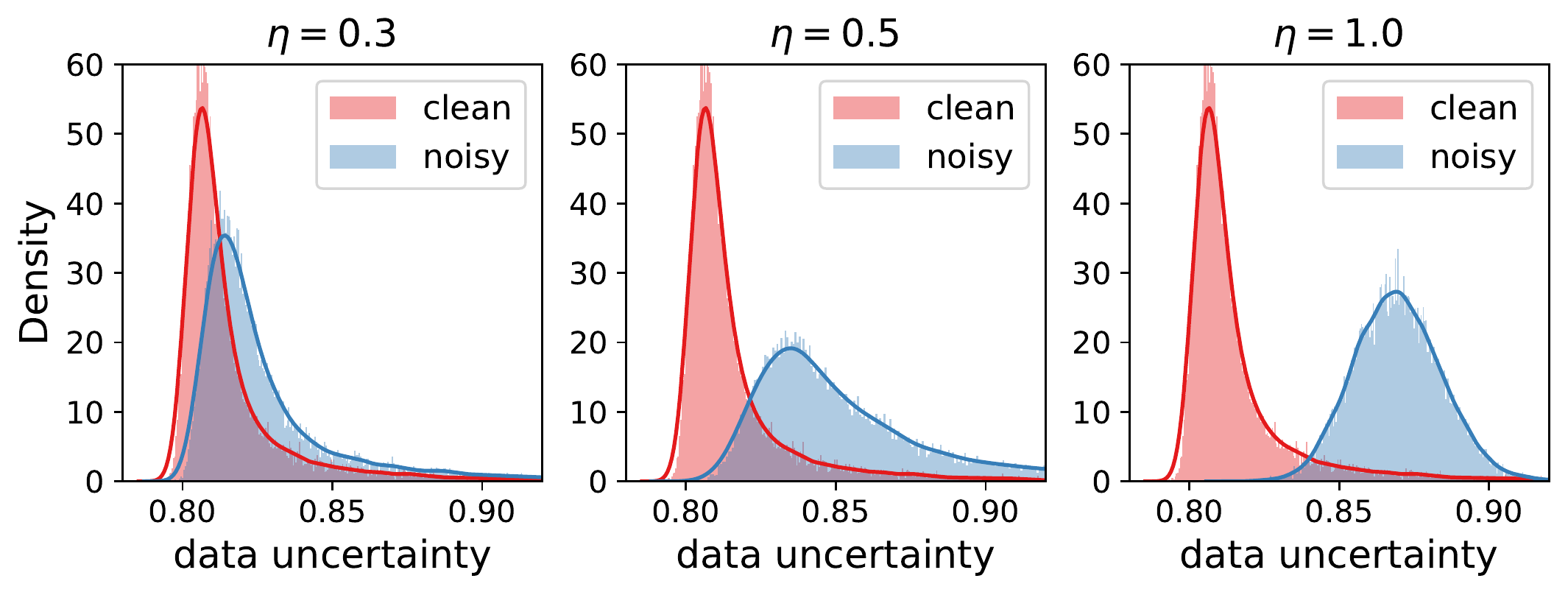}
\caption{ The role of data uncertainty in reliability assessment.  
It can be seen that samples with lower quality (larger noise) have larger data uncertainty.
}
\label{fig:data_uncertainty_capture}
\end{minipage}
\hfill
\begin{minipage}[h]{0.49\textwidth}
	\includegraphics[width=1.0\textwidth]{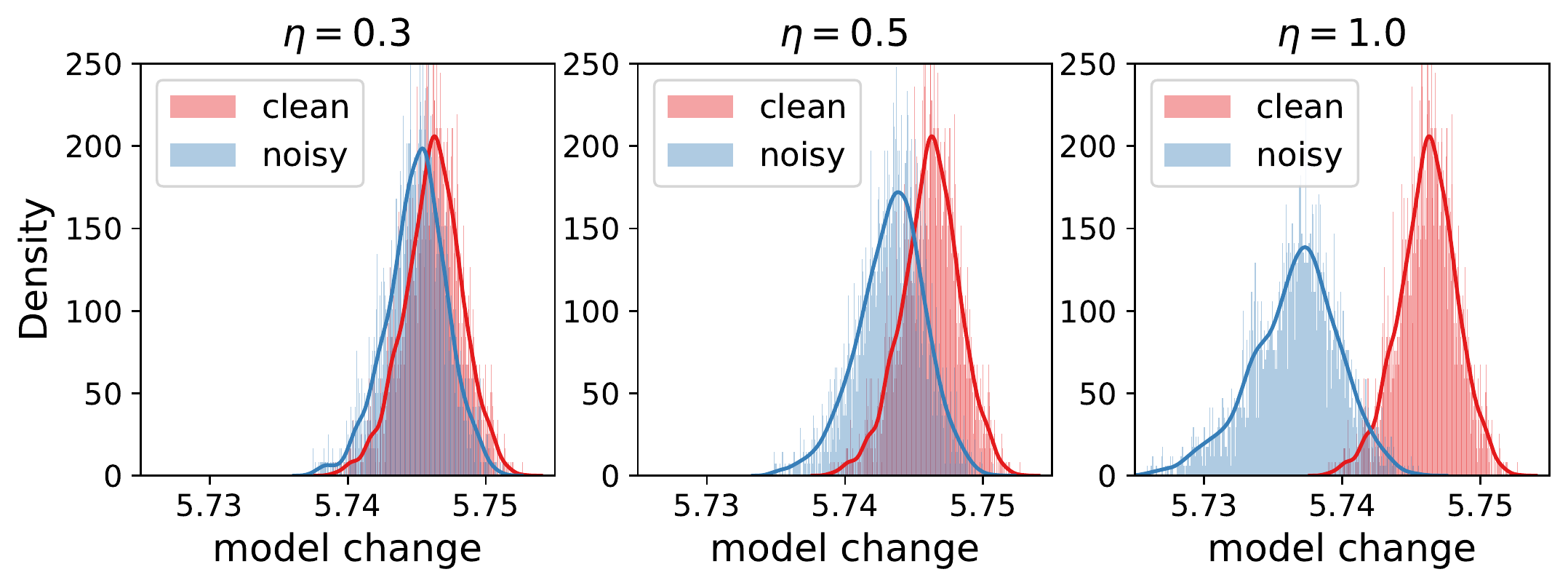}
	\caption{ The role of data uncertainty in the training process.
	It can be seen that samples with lower quality (larger noise) cause less model change in the training process. 
	}
	\label{fig:data_uncertainty_work}
\end{minipage}
\end{figure}

\textbf{The role of data uncertainty in reliability assessment.} We design experiments to verify whether the estimated data uncertainty can capture the ``noise" inherent in the data. 
We first regard samples in the gallery set of Market-1501~\cite{market} as clean data, and then add noise to pollute them to generate the noisy (low-quality) ones. Specifically, for an image tensor, \ie, $\boldsymbol{x}$, we generate a noise tensor $\boldsymbol{\varepsilon}$, where $ \boldsymbol{\varepsilon}\sim\mathcal{N}(\mathbf{0}, \mathbf{I})$. Then we pollute the origin data by $\boldsymbol{x} \leftarrow \boldsymbol{x} + \eta \boldsymbol{\varepsilon}$, where $\eta$ control the strength of the pollution. We gradually vary the size of $\eta$ to see how data uncertainty changes. The Gaussian kernel density estimation~\cite{scott2015multivariate} of the estimated data uncertainty are shown in Fig.~\ref{fig:data_uncertainty_capture}. When $\eta$ increases, the data uncertainty of noisy samples grows correspondingly. This shows that the proposed data uncertainty can capture the quality of the sample. 

\textbf{The role of data uncertainty in the training process.} We regard the samples in the training set of Market-1501 as clean samples and then pollute them to generate the noisy ones. We investigate how the clean and noisy samples affect model learning. Specifically, for each sample, we use it to individually train the model for 10 iterations, and count the model change. The model change is defined as the mean of the absolute difference of the model parameters before and after training. This can reflect the impact of the sample on the learning process. As shown in Fig.~\ref{fig:data_uncertainty_work}, when $\eta$ increases, the model change caused by noisy samples reduces, showing the proposed sampling-free data uncertainty learning method can suppress the contribution of low-quality samples during training.

\subsection{Comparison with State-of-the Art Methods}
\label{subsec: single query settings}

We also compare our method with state-of-the-art methods under the single-query settings. As shown in Table~\ref{table:results on holistic datasets.}, the compared methods are divided into two categories. One category employs ResNet50~\cite{ResNet} or slightly modifies ResNet50 without changing the main structure of the network. The other category employs more powerful backbones than ResNet50, such as HRNet-W32~\cite{hrnet} and Transformer~\cite{transformer}. For fair comparisons, we adopt ResNet50 and HRNet-W32 as our backbone, respectively. From the results, we can make several observations. 
(1) Whether using ResNet50 or HRNet-W32 as the backbone, our method consistently achieves comparable or superior performance on these datasets. Specifically, when HRNet-W32 is used as the backbone, our method outperforms the previous state-of-the-art methods by +4.5\% and +4.0\% in terms of mAP scores on MSMT17 and CUHK03-NP (detected), respectively. 
(2) Our method and ISP~\cite{ISP} have the same backbone and our method is much more effective. Especially, on CUHK03-NP (detected), our method outperforms ISP by +7.1\% mAP score. 
(3) Compared with semantic-based methods, \eg, FPR~\cite{FPR} and HOReID~\cite{HOReID}, our method does not need any additional external cues. Compared with methods that use more complex network structures, such as PAT~\cite{PAT} using the transformer~\cite{transformer}, our method still shows promising performance. 

\begin{table}[t]
\begin{center}
\caption{Comparison with state-of-the-art methods under the single-query settings.}
\label{table:results on holistic datasets.}
\begin{tabular}{l|c|cc|cc|cc|cc}
\toprule
\multirow{3}{*}{Methods} & \multirow{3}{*}{Backbone}  & \multicolumn{2}{c|}{\multirow{2}{*}{Market}} & \multicolumn{2}{c|}{\multirow{2}{*}{MSMT17}} & \multicolumn{4}{c}{CUHK03-NP} \\ \cline{7-10}
                  &         &       &         &  &  & \multicolumn{2}{c|}{labeled} &\multicolumn{2}{c}{detected} \\ \cline{3-10}
                  &       & R1     & mAP         & R1   & mAP     & R1     & mAP  & R1   & mAP \\ \hline
PCB+RPP~\cite{PCB}  & ResNet50     & 93.8      & 81.6      & --       & --          & 63.7   &   57.5 & --   & --           \\  
MGN~\cite{MGN}      & ResNet50     & \textbf{95.7}      & 86.9         & --       & --      & 68.0    &  67.4   & 66.8   &   66.0            \\  
CAMA~\cite{CAMA}    & ResNet50       &94.7          & 84.5              & --       & --      & 70.1     &  66.5   & 66.6    & 64.2                \\  
MHN-6~\cite{MHN-6}  & ResNet50       &95.1          & 85.0              & --       & --      & 77.2     &  72.4   & 71.7    & 65.4                \\  
FPR~\cite{FPR}      & ResNet50   &95.4          & 86.6               & --       & --      & 76.1     & 72.3    & --      & --          \\ 
HOReID~\cite{HOReID} & ResNet50        & 94.2         & 84.9                 & --       & --      & --       & --      & --      & --                 \\  
DGNet~\cite{DG-Net}  & ResNet50    & 94.8   & 86.0       & 77.2     & 52.3     & --  & --  & -- & --   \\ \hline
\textbf{UAL}\textit{(Ours)} & ResNet50  & 95.2   & \textbf{87.0}        & \textbf{80.0}    &\textbf{56.5}    & \textbf{78.2}    & \textbf{75.6}    & \textbf{76.1}     &\textbf{72.0}  \\  \hline \hline
RGA-SC~\cite{RGA-SC} & R50-RGA      & \textbf{96.1}     & 88.4      & 80.3 & 57.5 & 81.1     & 77.4    & 79.6    & 74.5            \\  
ABD-Net~\cite{ABD-Net} & ABD-Net  & 95.6     & 88.3        & 82.3     & 60.8    & --       &  --     & --      & --                 \\  
BAT-net~\cite{BAT-net} & BAT-net        &95.1       & 87.4       & 79.5     & 56.8    & 78.6     &  76.1   & 76.2    & 73.2               \\   
OSNet~\cite{OSNet}  & OSNet  & 94.8          & 84.9          & 78.7     & 52.9    & --       & --    & 72.3      & 67.8                 \\   
ISP~\cite{ISP}      &  HRNet   & 95.3          & 88.6       & --       & --      & 76.5     & 74.1    & 75.2    & 71.4          \\ 
PAT~\cite{PAT}      &  Transformer  &95.4          & 88.0            & --       & --      & --       & --      & --      & --            \\  \hline 
\textbf{UAL}(\textit{Ours})       & HRNet  & 95.7 & \textbf{89.5}   &  \textbf{84.7} & \textbf{65.3}    &\textbf{83.7} &\textbf{81.0} & \textbf{81.0} &\textbf{78.5} \\  
\bottomrule
\end{tabular}
\end{center}
\end{table}

\subsection{Ablation Study}
\label{sec:ablation}
In this part, we conduct ablation studies to show the effectiveness of each component of the proposed method.

\textbf{The effectiveness of data uncertainty and model uncertainty.} 
Besides based on the backbones ResNet50 and HRNet-W32 implemented by ours, we also conduct the experiments based on the more in-data baseline in \textbf{fast-reid}, \ie, BOT~\cite{BOT}, with different backbones including ResNet50 (R50), a variant with IBN layers (R50-ibn) and ResNeSt (S50). Experiments are conducted on MSMT17~\cite{MSMT17} dataset. As shown in Table~\ref{table:ablation of aleatoric and epistemic uncertainty.}, both learning data uncertainty and learning model uncertainty improve the performance, and learning data uncertainty provides a larger improvement.
The results show these two uncertainties can provide complementary information for learning discriminative latent space.

\textbf{Impact of hyper-parameter $T$.} During testing, for a sample $\boldsymbol{x}$, we need to sample $T$ times from $q_{\pi}(\boldsymbol{\theta})$ to obtain the $\bar{\boldsymbol{\mu}}$ and $\sigma^2_{m}$ (as described in Sec.~\ref{subsec: a unified network}). Here, we show how $T$ affects the performance on CUHK03-NP~\cite{CUHK03} and Market-1501~\cite{market}. The results are shown in Table~\ref{table:ablation of T.}. 
We also report the time cost of the entire testing process, from extracting features to calculating the Rank-1 scores.
As we can see, our method can work well with a small $T$. Further, our method outperforms HOReID~\cite{HOReID} and PGFA~\cite{PGFA} by a large margin while the time cost is much smaller, which shows the effectiveness and efficiency of our method. 

\begin{table}[t]
\begin{center}

\caption{Effectiveness of data uncertainty and model uncertainty. Experiments are conducted on MSMT17~\cite{MSMT17}. $^\dag$ indicates the experiments are conducted on the backbones in publicly available repository fast-reid: \textcolor{magenta}{\url{\textbf{https://github.com/JDAI-CV/fast-reid}}}}
\label{table:ablation of aleatoric and epistemic uncertainty.}
\begin{tabular}{cc|cc|cc|cc|cc|cc}
\hline
 \multicolumn{2}{c|}{Uncertainty} & \multicolumn{2}{c|}{ResNet50} & \multicolumn{2}{c|}{$^\dag$BOT(R50)} & \multicolumn{2}{c|}{$^\dag$BOT(R50-ibn)} & \multicolumn{2}{c|}{$^\dag$BOT(S50)}  & \multicolumn{2}{c}{HRNet-W32} \\
Data & Model & Rank-1 & mAP  & Rank-1  & mAP  & Rank-1  & mAP  & Rank-1  & mAP  & Rank-1  & mAP \\ \hline
 $\times$ & $\times$ & 76.5 & 52.0 & 73.9  & 49.9 & 79.1 & 55.4 & 81.0 & 59.4 & 81.6  & 59.6 \\
 \checkmark & $\times$  & 77.7  & 53.1 & 77.1 & 51.4 & 80.4 & 56.0 & 81.9 & 59.5 & 83.4 & 62.2 \\
 $\times$   & \checkmark  & 76.7 & 52.3 & 76.6 & 51.2 & 80.8 & 56.9 & 82.6 & 60.8 &  81.9  & 60.0  \\ 
 \checkmark & \checkmark  &\textbf{80.0} & \textbf{56.5} & \textbf{78.7} & \textbf{53.6} & \textbf{82.4} & \textbf{59.1} & \textbf{84.1} & \textbf{62.1} &\textbf{84.7} & \textbf{65.3} \\
\hline
\end{tabular}
\end{center}
\end{table}

\begin{table}[t]
\begin{center}
\caption{Impact of the hyper-parameter $T$.
We use the AMD EPYC 7742 CPU and GeForce RTX 3090 GPU.
The backbone is ResNet50.}
\label{table:ablation of T.}
\begin{tabular}{c|ccc|ccc}
\hline
\multirow{2}{*}{$T$}  &  \multicolumn{3}{c|}{CUHK03 (Labeled)} & \multicolumn{3}{c}{Market-1501} \\ \cline{2-7} 
      & Time  &  R1 & mAP & Time & R1 & mAP \\ \hline
HOReID~\cite{HOReID} & --    & --   & --  & 236 s   & 94.2    & 84.9 \\ 
PGFA~\cite{PGFA} & -- & --  & --   & 193 s   & 91.2     & 76.8 \\  \hline  
 Ours ($T$=5) & 17 s & 78.4  & 75.5  & 66 s  & 95.0  & 86.9\\    
 Ours ($T$=10) & 24 s & \textbf{78.6} & \textbf{75.6} & 87 s & \textbf{95.2} & \textbf{87.0} \\ 
 Ours ($T$=20) & 39 s & 78.3 & \textbf{75.6} & 138 s & 95.1  & \textbf{87.0} \\
\hline
\end{tabular}
\end{center}
\end{table}

\section{Conclusions}
In this paper, we propose an Uncertainty-Aware Learning (UAL) method for the ReID task to provide reliability-aware predictions, which is achieved by considering two types of uncertainty: data uncertainty and model uncertainty. These two types of uncertainty are integrated into a unified network for joint learning without any external clues. Comprehensive experiments under the risk-controlled settings and the multi-query settings verify that the proposed reliability assessment is effective. Our method also shows superior performance under the single query settings. Meanwhile, we also provide quantitative analyses of the learned data uncertainty and model uncertainty. We expect that our method will provide new insights and attract more interest in the reliability issue in person ReID.

\textbf{Acknowledgement.} This work was supported by the state key development program in 14th Five-Year under Grant Nos. 2021YFF0602103, 2021YFF0602102, 2021QY1702. We also thank for the research fund under Grant No. 2019GQG0001 from the Institute for Guo Qiang, Tsinghua University.

\clearpage
%
%


\bibliographystyle{splncs04}
\bibliography{egbib}
\end{document}